\documentclass[journal]{IEEEtran}

\ifCLASSINFOpdf
\else
   \usepackage[dvips]{graphicx}
\fi
\usepackage{url}

\usepackage{graphicx}
\usepackage{amsmath}
\usepackage{caption}
\usepackage{amssymb}
\usepackage{float} 
\usepackage{subcaption}
\usepackage{tabularx}
\usepackage{xstring}
\usepackage{multirow}

\begin{document}

\title{Data Augmentation Vision Transformer for Fine-grained Image Classification}

\author{Chao Hu, \IEEEmembership{Member, IEEE}, Liqiang Zhu, Weibin Qiu, and Weijie Wu}

\markboth{IEEE SIGNAL PROCESSING LETTERS, VOL. 28, 2022}
{Shell \MakeLowercase{\textit{et al.}}: Bare Demo of IEEEtran.cls for IEEE Journals}
\maketitle

\begin{abstract}
Recently, the vision transformer (ViT) has made breakthroughs in image recognition. Its self-attention mechanism (MSA) can extract discriminative labeling information of different pixel blocks to improve image classification accuracy. However, the classification marks in their deep layers tend to ignore local features between layers. In addition, the embedding layer will be fixed-size pixel blocks. Input network Inevitably introduces additional image noise. To this end, we study a data augmentation vision transformer (DAVT) based on data augmentation and proposes a data augmentation method for attention cropping, which uses attention weights as the guide to crop images and improve the ability of the network to learn critical features. Secondly, we also propose a hierarchical attention selection (HAS) method, which improves the ability of discriminative markers between levels of learning by filtering and fusing labels between levels.  Experimental results show that the accuracy of this method on the two general datasets, CUB-200-2011, and Stanford Dogs, is better than the existing mainstream methods, and its accuracy is 1.4\% and 1.6\% higher than the original ViT, respectively
\end{abstract}

\begin{IEEEkeywords}
fine-grained visual classification, data augmentation, hierarchical attention selection, self-attention mechanism.
\end{IEEEkeywords}

\IEEEpeerreviewmaketitle

\section{Introduction}
\IEEEPARstart{U}{nlike} traditional image classification, fine-grained vision classification (FGVC) refers to classifying subordinate categories into basic categories, for example bird and dog species. Due to significant within-class differences and low between-class variance, objects in the same category often exhibit completely different morphologies, and objects of different classes may also be very similar. Second, due to the limitations of training data, fine-grained labeling categories often requires a great deal of expertise, so fine-grained image recognition is considered a more challenging task.

Most previous work \cite{ref1}-\cite{ref4} has used convolutional neural networks (CNNs) as the backbone network to extract subtle features in images. However, with the deepening of the number of network layers, the computational complexity of such methods increases, and it is easily disturbed by non-feature region noise when extracting deep features. Currently, the mainstream fine-grained classification methods include positioning-based \cite{ref2}-\cite{ref5} and feature coding methods \cite{ref6}-\cite{ref9}. The former uses attention mechanisms, clustering, and other means to discover specific areas. The latter is to capture more subtle features of the region by calculating high-order information. The low labeling cost of weakly supervised learning lead that mainstream classification models have focused more on various attention-guided methods.

In recent years,  with the breakthrough progress \cite{ref10}-\cite{ref12} of Transformer in computer vision, Transformer has shown superior performance in general image classification \cite{ref10}, image retrieval \cite{ref13}, and semantic segmentation \cite{ref14}. The Visual Transformer (ViT) has proven its great potential in image classification by automatically identifying discriminating feature areas in images through its inherent attention mechanism. However, ViT's perception can't be effectively extended, and the length of its pixel block markers doesn't change as the number of encoder layers increases, which lead ViT doesn't work on FGVC. In addition, ViT input fixed-size blocks of pixels are not conducive to the network capturing critical area attention information. To address the potential problems with ViT, AFTrans \cite{ref15} captures regional attention through an adaptive attention multiscale fusion transformer. A peak suppression module \cite{ref16} penalizes the most discriminating region to learn different fine-grained representations. TransFG \cite{ref17} and FFVT \cite{ref18} scale ViT to large-scale and small-scale FGVC datasets, respectively. However, these efforts ignore the noise effects of some local features between levels and fixed-size pixel blocks. Inspired by the FFVT model, we extend critical marker filtering to all levels of the ViT model. It uses the cross-multiplication method of levels to extract local features layer by layer, emphasizing the tag correlation between levels. Meanwhile, the image enhancement method based on the attention mechanism is extended to Transformer, and the effect of feature enhancement is achieved by using data cropping to erase irrelevant image noise.

This letter proposes a novel data-augmentation visual transformer (DAVT) for fine-grained image classification. It aggregates and classifies discriminating feature information at all levels. Its hierarchical attention selection HAS compensates for missing information between parts and levels by filtering and merging key markers across levels. Secondly, attention-guided image cropping is used to reduce some of the noise interference caused by fixed-size pixel blocks and enhance the ability of images to express critical characteristics.

\begin{figure}
	\centering
    \includegraphics[width=0.48\textwidth]{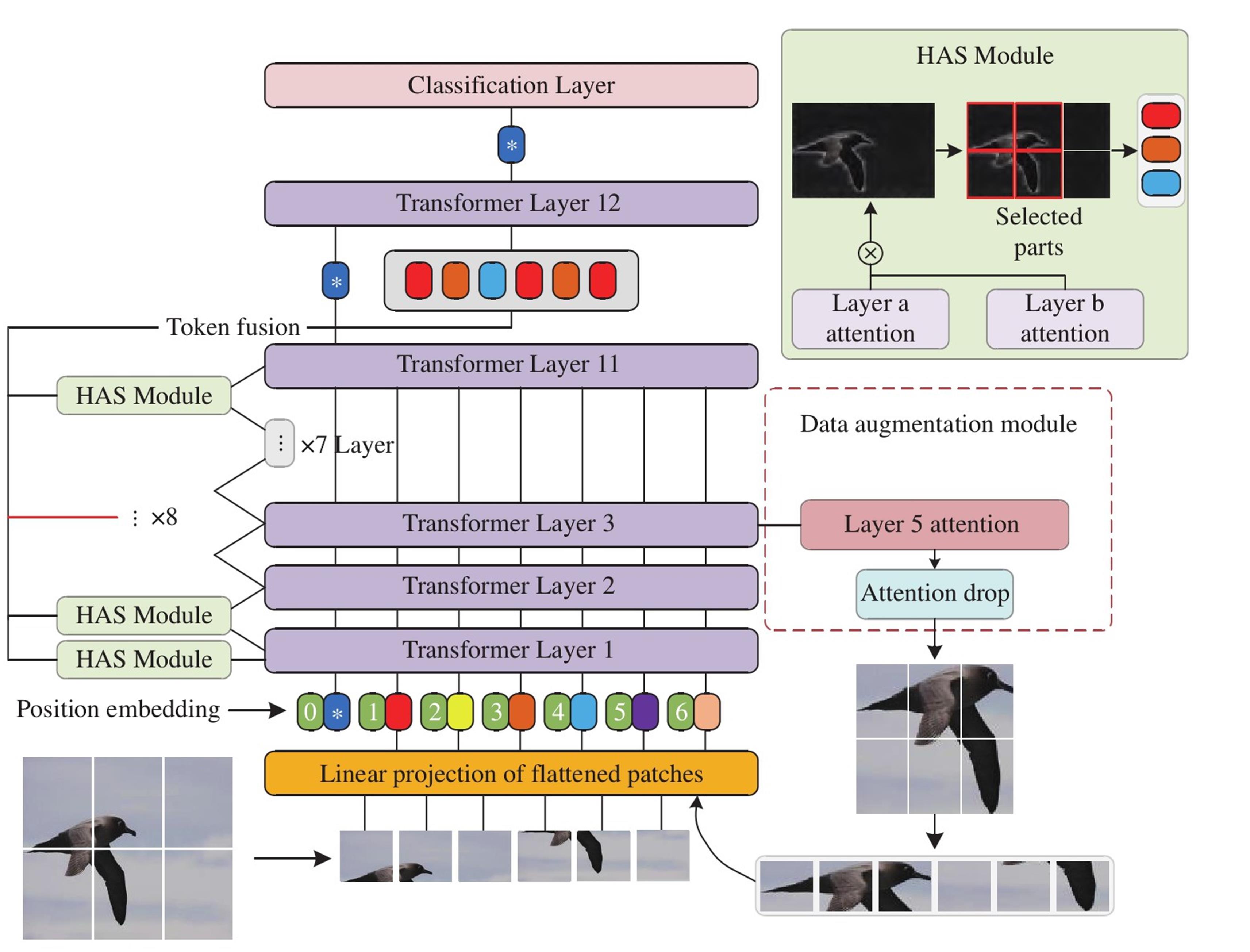}
	\caption{Overall Architecture of DAVT}
	\label{fig:1}
\end{figure}

\section{Proposed Method}
Because the input markers deep in the ViT network model are difficult to focus on the subtle features of the image part, it is also challenging to aggregate important marker information between the layers. we propose a hierarchical attention selection and data augmentation method to force the network to select discriminative marker information to compensate for missing feature information between various levels. The augmentation image is looped from the new input network to enable the network to generate multi-scale features to learn the discriminant region. The overall structure of the DAVT is shown in Fig. \ref{fig:1}.

\subsection{Hierarchical Attention Selection Mechanism}
The marker information of the embedding layer in the original ViT is less discriminative, and the high-level input tag may not necessarily retain the relatively important feature information of the original mark due to the noise caused by the fixed-size pixel block. Therefore, we propose a hierarchical attention selection mechanism, which fuses the attention weights between levels and screens the markers. The attention weights of the first few layers in the DAVT network model are expressed as:
\begin{equation}
\label{equation:1}
\begin{aligned}
a_l&=[a^0_l,a_l^1,a_l^2,\cdots,a_l^K]\qquad l\in 1,2,\cdots,L-1 \\
a_l^i&=[a^{i_0}_l,a_l^{i_1},a_l^{i_2},\cdots,a_K^{i_N}]\qquad i\in 1,2,\cdots,K
\end{aligned}
\end{equation}

\noindent where $K$ represents the number of self-attention heads in the multi-head self-attention mechanism in the model. $l$ represents the number of layers of the model but only relies on the self-attention head and tail to extract marker information within each level, which is easy to cause the omission of some local information between levels in order to utilize the marker information between each level fully, the attention weight between levels is integrated in the following way:
\begin{equation}
\label{equation:2}
\begin{aligned}
h_la=a_l*a_{l+1}\qquad l\in 1,2,\cdots,L-1
\end{aligned}
\end{equation}

\noindent As shown in Eq. \ref{equation:2}, $h_la$ represents the attention weight after $l$ layer fusion. Specifically, the attention weights of the $l$ and $l+1$ layers are fused by matrix multiplication, where $l\in(0,L-1)$. Apply this method to all Transformer layers that remove the last layer. The $Max()$ function is then used to select the maximum indicators $A_1, A_2,\cdots,A_K$ for $K$ different attention heads, which are used as indexes of the model to extract discriminant markers between levels. The key tags selected using the HAS method are represented as:
\begin{equation}
\label{equation:3}
z_l^{local}=[z_l^{A_1},z_l^{A_2},\cdots,z_l^{A_K}]
\end{equation}
\noindent where $A_K$ represents the number of attention-weight selected features. Finally, the classification marks are connected to the discriminative marks filtered by each layer, and after fusion, they are input into the last Transformer layer for classification.
\begin{equation}
\label{equation:4}
z_f=[z^0_{L-1};z_1^{local},z_2^{local},\cdots,z_{L-1}^{local}]
\end{equation}

The hierarchical attention selection method directly aggregates discriminative markers between levels. It connects categorical and fused markers as input to the last transformer layer, which preserves local information between levels and fully uses inter-level attention weight to capture local information lost in the depths.

\begin{table*}[!t]
\renewcommand{\arraystretch}{1.2}
\caption{\label{tab1}\centering{Ablation studies under the CUB-200-2011 and Stanford Dogs datasets}}
\centering
\setlength{\tabcolsep}{15pt}
{\begin{tabular}{cccccc}
\hline
Method  &   CUB-200-2011/Top-1\%    &   Stanford Dogs/Top-1\%   &   Time/h  &   Parameters  &   FLOPs   \\
\hline
ViT \cite{ref5}     &   90.4    &   89.8    &   9   &   $8.67\times10^7$  &   $17.61\times10^9$     \\
ViT+HAS     &   91.5    &   90.5    &   9   &   $8.91\times10^7$  &   $17.72\times10^9$     \\
DAVT     &   91.8    &   91.0    &   11.7   &   $10.87\times10^7$  &   $19.31\times10^9$     \\
\hline
\end{tabular}}
\end{table*}

\subsection{Attention Guide Data Augmentation}
During training, ViT splits the original image into fixed-size pixel blocks, and as the amount of data increases, the image noise introduced by the pixel blocks also increases. For example, in a specific pixel block, there is no crucial information such as the bird's head, wings, neck, etc., only the background information of the environment, such as sky and clouds. Attention-guided data enhancement can weaken the effect of image noise and increase the data in the training part.

During training, the process of array enhancement is guided by the attention map $\alpha_\xi$ of the $\xi$ layer generated by the multi-head self-attention mechanism in ViT, and normalized to $\alpha^{*}_\xi$ to make it more representative. The normalization process is as follows:
\begin{equation}
\label{equation:5}
\alpha^{*}_\xi=\frac{\alpha_\xi-\min(\alpha_\xi)}{\max(\alpha_\xi)-\min(\alpha_\xi)}
\end{equation}

\noindent where $\alpha^{*}_\xi$ is the normalized attention map, and subsequent experiments show that the model's performance is optimal when $\xi=5$.

This method uses the augmentation attention map $\alpha^{*}_\xi$ to crop the attention of the image, zoom in on the area of the part of the object, and extract more detailed local features. Specifically, the clipping mask $C_k$ is obtained by normalized attention map $\alpha^{*}_\xi$ sets elements $\alpha^{*}_\xi(i,j)$ to bindify it, and the root binary mask sets the value of elements greater than the threshold $\theta_c$ to $1$ and the other elements to $0$, where $\theta_c\in[0.4,0.6]$. Elements with a value of $1$ represent discriminant regions in the image, while elements with $0$ represent background noise. The specific process is expressed as
\begin{equation}
\label{equation:6}
C_k(i,j)=\begin{cases}1,    &   if\quad \alpha^{*}_\xi(i,j)>\theta_c   \\
0,  &   otherwise
\end{cases} 
\end{equation}

The goal is to find the smallest bounding box that can cover the entire Mask (element value $1$) selected by $C_k$, crop the original image with this bounding box, and zoom in to the original image size as the enhanced input data. Due to the increased proportion of the object part, the model can better locate the object position and facilitate the extraction of more discriminative features.

Due to the introduction of new images after data augmentation, the loss function consists of several parts during the training phase. As shown in Eq. \ref{equation:7}:
\begin{equation}
\label{equation:7}
Loss_{total}=Loss_V+Loss_C
\end{equation}
\noindent where$Loss_V$ and $Loss_C$ represent the cross-entropy loss of the original image and the cross-entropy loss after attention cropping, respectively. These two losses work together during backpropagation to optimize the model's performance.

\begin{table}[!t]
\renewcommand{\arraystretch}{1.2}
\caption{\label{tab2}\centering{Ablation experiment of attention extraction layer $\xi$ under CUB-200-2011 dataset}}
\centering
\setlength{\tabcolsep}{21.8pt}
{\begin{tabular}{ccc}
\hline
Method  &  $\xi$    &   CUB-200-2011/Top-1\%  \\
\hline
DAVT    &   1   &   91.4    \\
DAVT    &   3   &   91.5    \\
DAVT    &   5   &   91.8    \\
DAVT    &   7   &   91.6    \\
DAVT    &   9   &   91.6    \\
DAVT    &   11   &   91.5    \\
\hline
\end{tabular}}
\end{table}

\begin{table}[!t]
\renewcommand{\arraystretch}{1.2}
\caption{\label{tab3}\centering{Comparison of different methods under the CUB-200-2011 and Stanford Dogs datasets}}
\centering
\resizebox{\linewidth}{!}
{\begin{tabular}{ccccc}
\hline
Method      &           Backbone     &   1-Stage    &       CUB-200-2011     &   Stanford Dogs   \\
\hline
ResNet50 \cite{ref21}    &        ResNet50        &   \checkmark    &        84.5        &           82.7    \\
GP-256 \cite{ref22}   &         VGG16        &   $\times$    &   85.8    &       83.1    \\
DFL-CNN \cite{ref23}   &       ResNet50        &   $\checkmark$    &   87.4    &       87.7    \\
NTS-Net \cite{ref24}   &       ResNet50        &   $\checkmark$    &   87.5    &       88.2    \\
Cross-X \cite{ref10}   &       ResNet50        &   $\times$    &   87.7    &       88.9    \\
CIN \cite{ref11}   &        ResNet101        &   $\checkmark$    &   88.1    &       87.6    \\
ACNet \cite{ref25}  &   ResNet50    &   $\checkmark$    &   88.1    &   87.4    \\
S3N \cite{ref8}    &   ResNet50    &   $\checkmark$    &   88.5    &   87.1    \\
FDL \cite{ref4}    &   DenseNet161     &   $\checkmark$    &   89.1    &   84.9    \\
FBSD \cite{ref3}    &   DenseNet161     &   $\checkmark$    &   89.8    &   88.1    \\
API-Net \cite{ref9}    &   DenseNet161     &       $\checkmark$    &   90.0    &   90.3    \\
CAL \cite{ref2}        &   ResNet101       &       $\checkmark$    &   90.4    &   88.7    \\
ViT \cite{ref5}       &   ViT-B\_16       &   $\checkmark$        &   90.4    &   89.4    \\
TransFG \cite{ref17}    &   ViT-B\_16       &   $\checkmark$        &   91.4    &   90.4    \\
FFVT \cite{ref18}        &   ViT-B\_16       &   $\checkmark$        &   91.4    &   90.4    \\
RAMS \cite{ref16}        &   ViT-B\_16       &   $\checkmark$        &   91.5    &   90.7    \\
AFTrans \cite{ref5}     &   ViT-B\_16       &   $\checkmark$        &   91.6    &   90.6    \\
DAVT        &   ViT-B\_16       &   $\checkmark$        &   91.8    &   91.0    \\
\hline
\end{tabular}}
\end{table}

\section{Experiments}
\subsection{Datasets}
This letter evaluates the model on CUB-200-2011 \cite{ref19}, the Stanford Dogs \cite{ref20} dataset. CUB-200-2011 is a bird dataset containing 11788 bird pictures, including 5994 pictures in the training and 5794 images in the test set, for a total of 200 bird categories. Stanford Dogs contains images of 120 dog species from around the world, including 12000 pictures in the training set and 8580 pictures in the test set. The two datasets contain both image labels, bounding boxes, and part notes. In the experiment, Top-1\% is used to represent the accuracy of the classification of fine-grained images, and the rate of change of the loss function is used to test whether the training process has overfitting.

\subsection{Experiment Settings}
In all experiments, the network was trained by loading the weights of the official ViT-B\_16 model pre-trained on ImageNet21k, resizing the original image to $448\times448$, and segmenting the image to $16\times16$ pixel blocks. The training period for each dataset is set to 10000. The experiment used random horizontal flipping to augment the image data. During the training phase, the network is optimized with a momentum value of 0.9 using stochastic gradient descent (SGD), the initial learning rate is set to 0.02, and the learning rate is orderly adjusted using cosine annealing. The batch size is set to 6. we use the Pytorch framework as an experimental platform, accelerated training in FP16 data format via the APEX toolkit. A single NVIDIA TESLA T4 graphics card  is uesd for experimentation. The value of hyperparameter $K$ is set to 12, and hyperparameter $\xi$ is set to 5.
\begin{figure}[!t]
	\centering
    \includegraphics[width=0.35\textwidth]{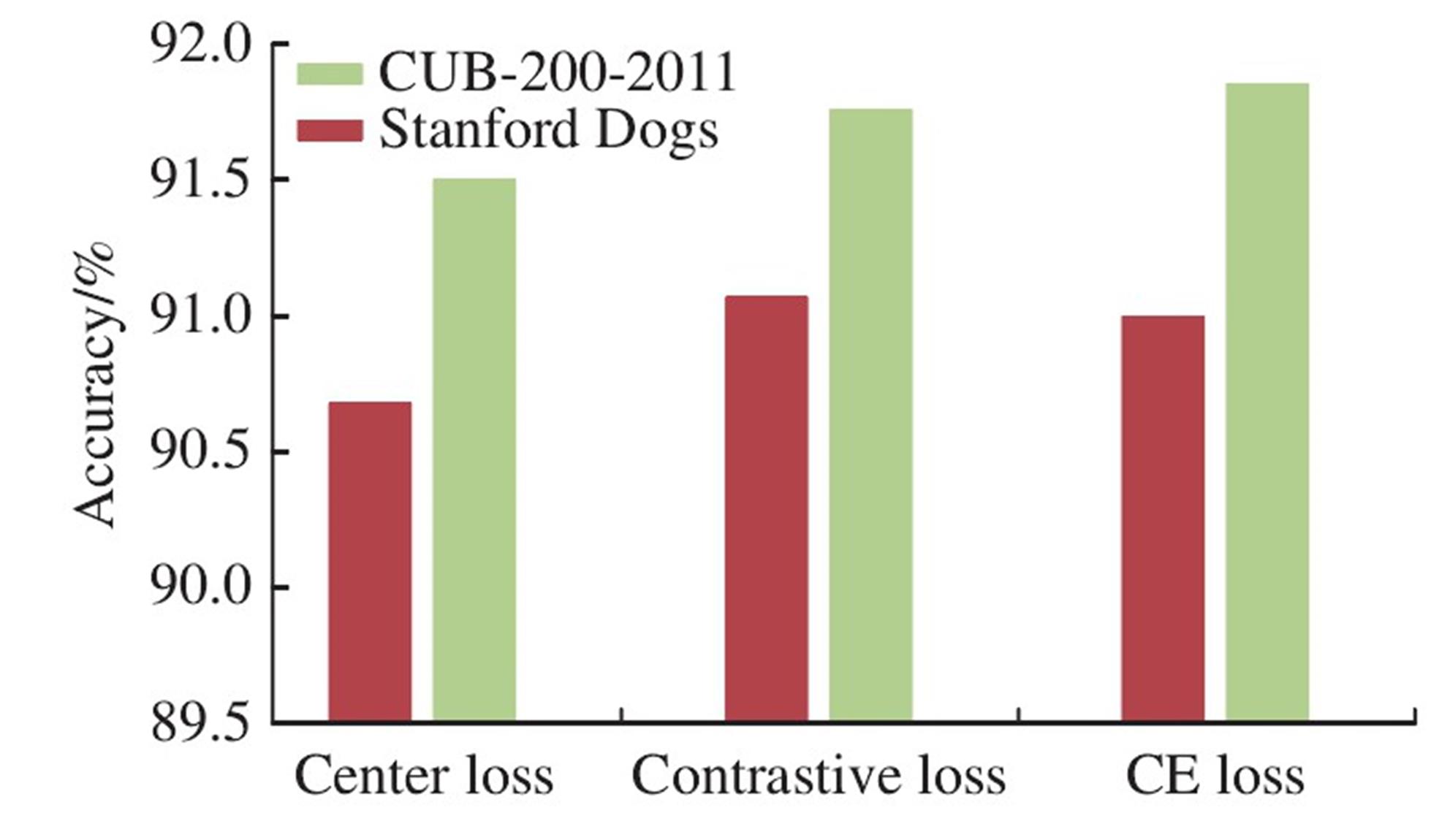}
	\caption{Different loss function}
	\label{fig:2}
\end{figure}

\begin{figure}[!t]
	\centering
    \includegraphics[width=0.4\textwidth]{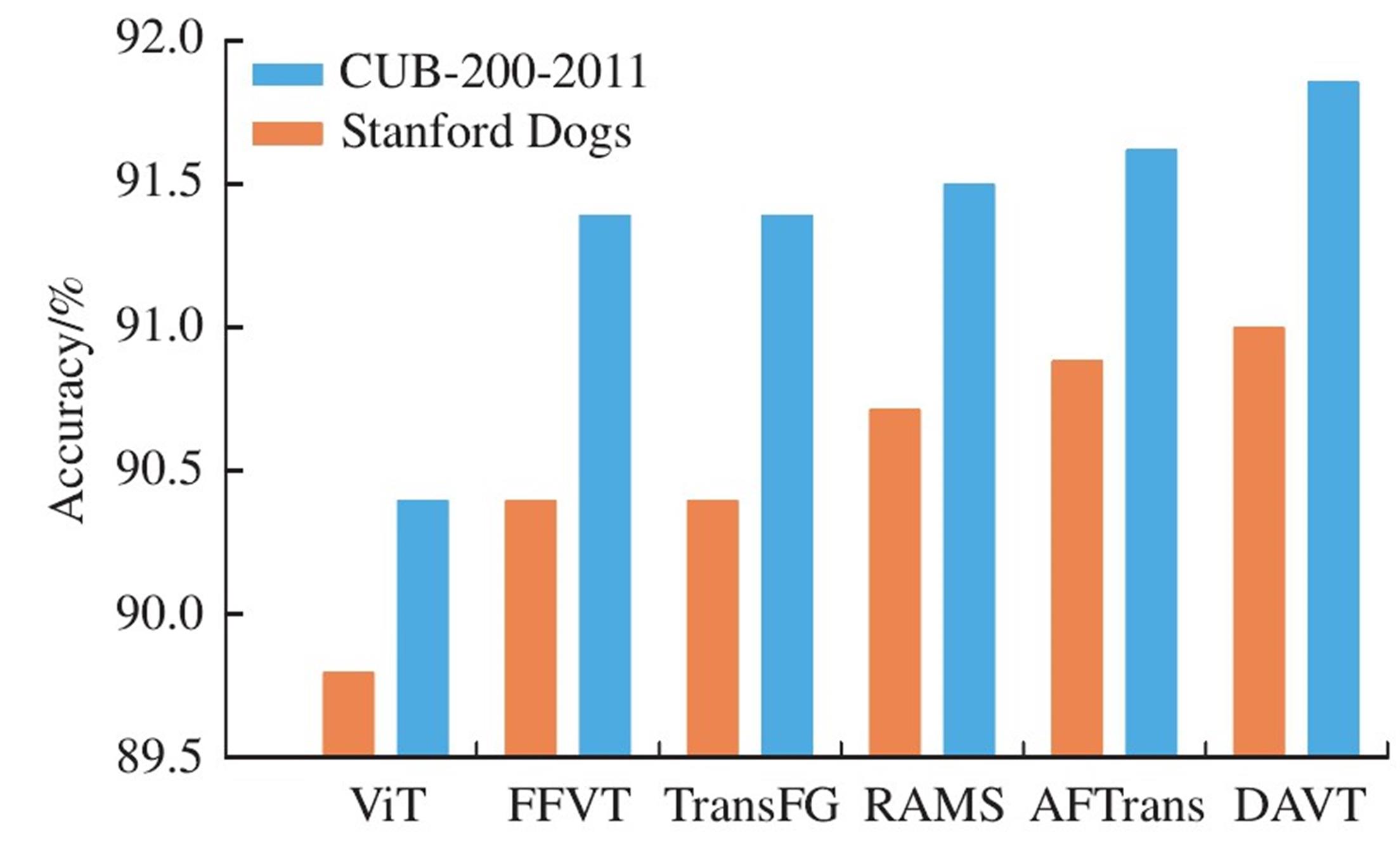}
	\caption{Performance comparison of state-of-the-art methods}
	\label{fig:3}
\end{figure}

\begin{figure*}
    \centering
    \begin{subfigure}[t]{0.225\textwidth}
           \centering
           \includegraphics[width=\textwidth]{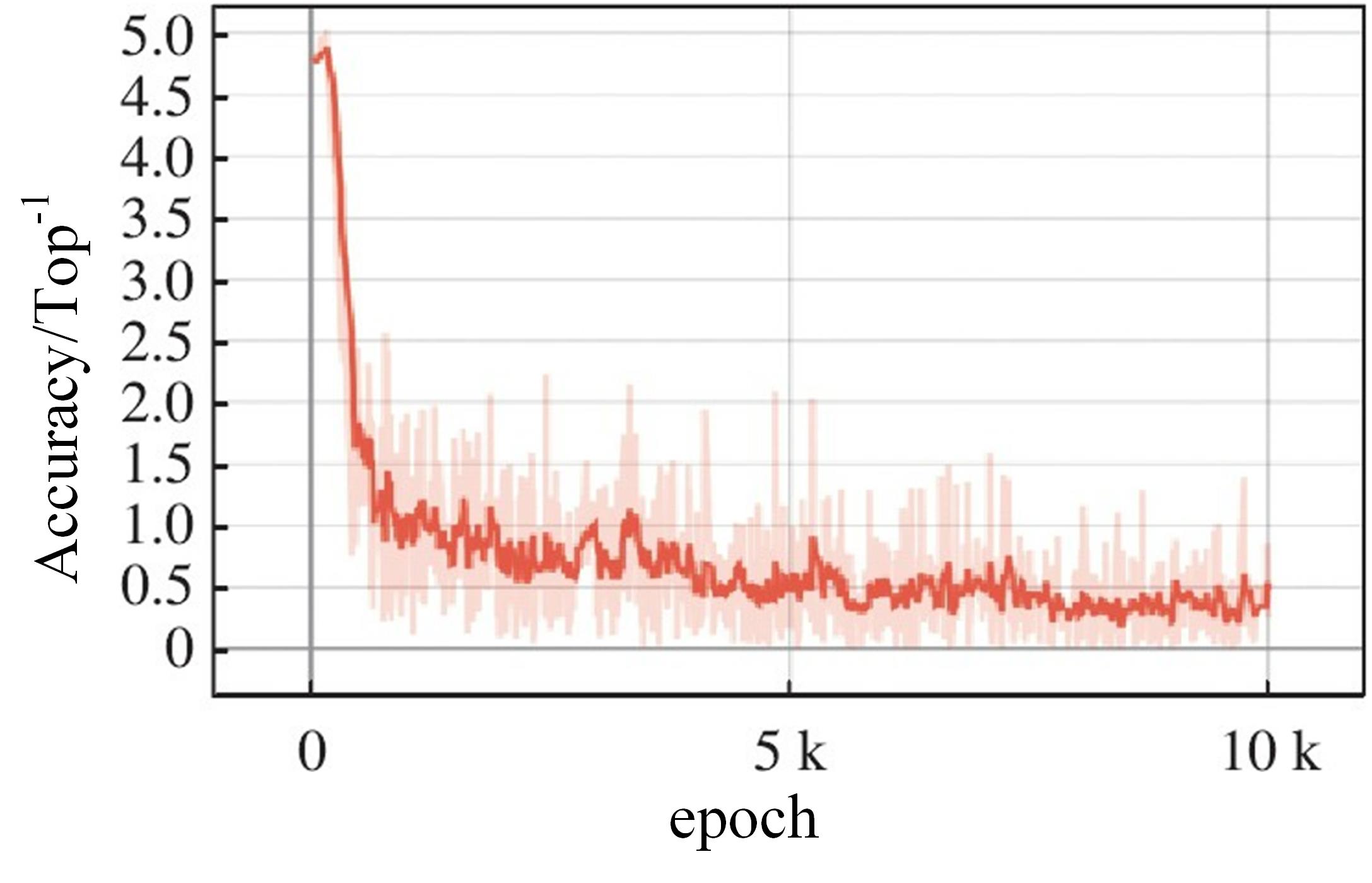}
           \caption{Stanford Dogs loss\label{fig4a}}
    \end{subfigure}
    \quad
    \begin{subfigure}[t]{0.225\textwidth}
            \centering
            \includegraphics[width=\textwidth]{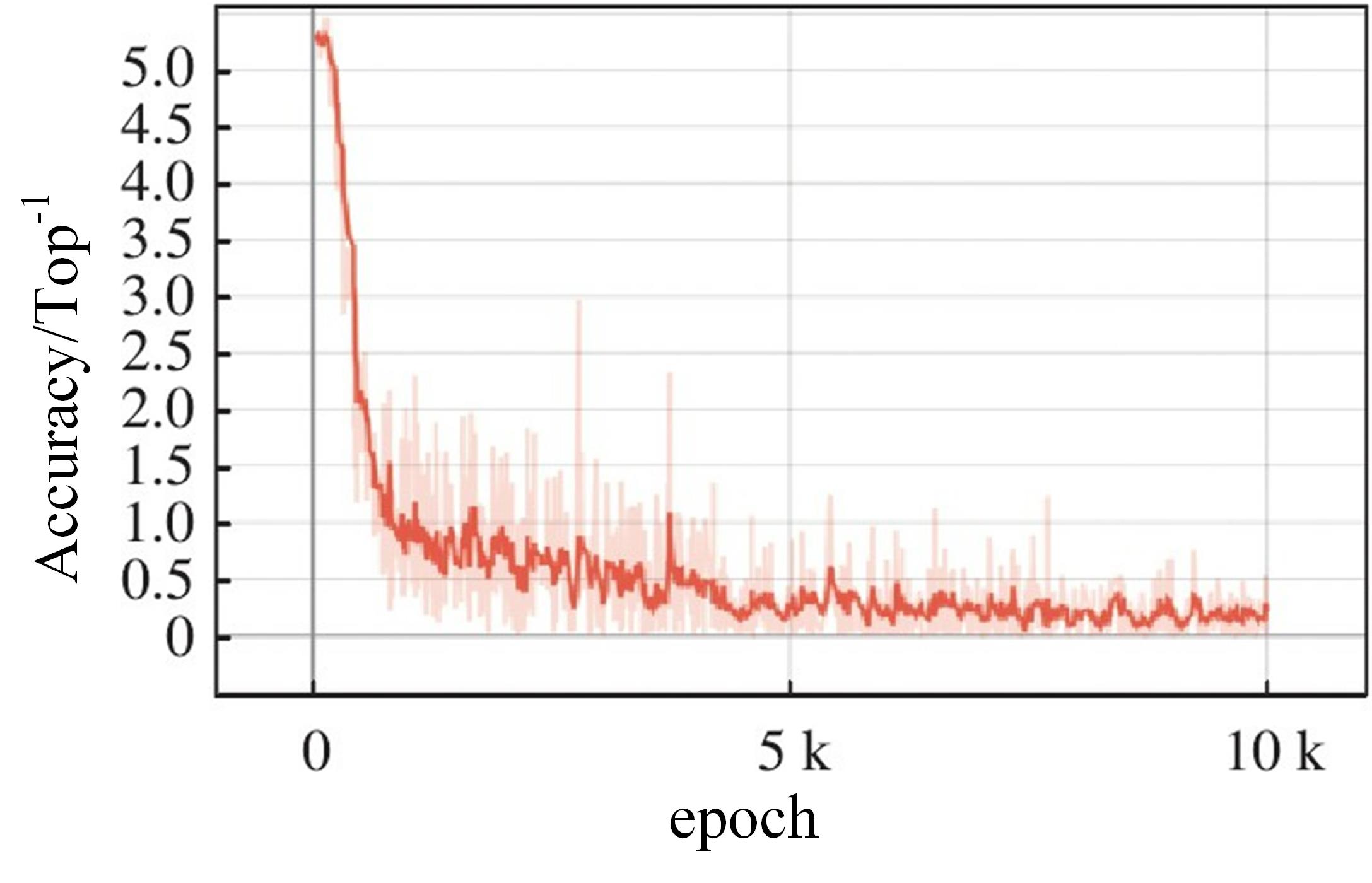}
            \caption{CUB-200-2011 loss\label{fig4b}}
    \end{subfigure}
    \quad
    \begin{subfigure}[t]{0.225\textwidth}
            \centering
            \includegraphics[width=\textwidth]{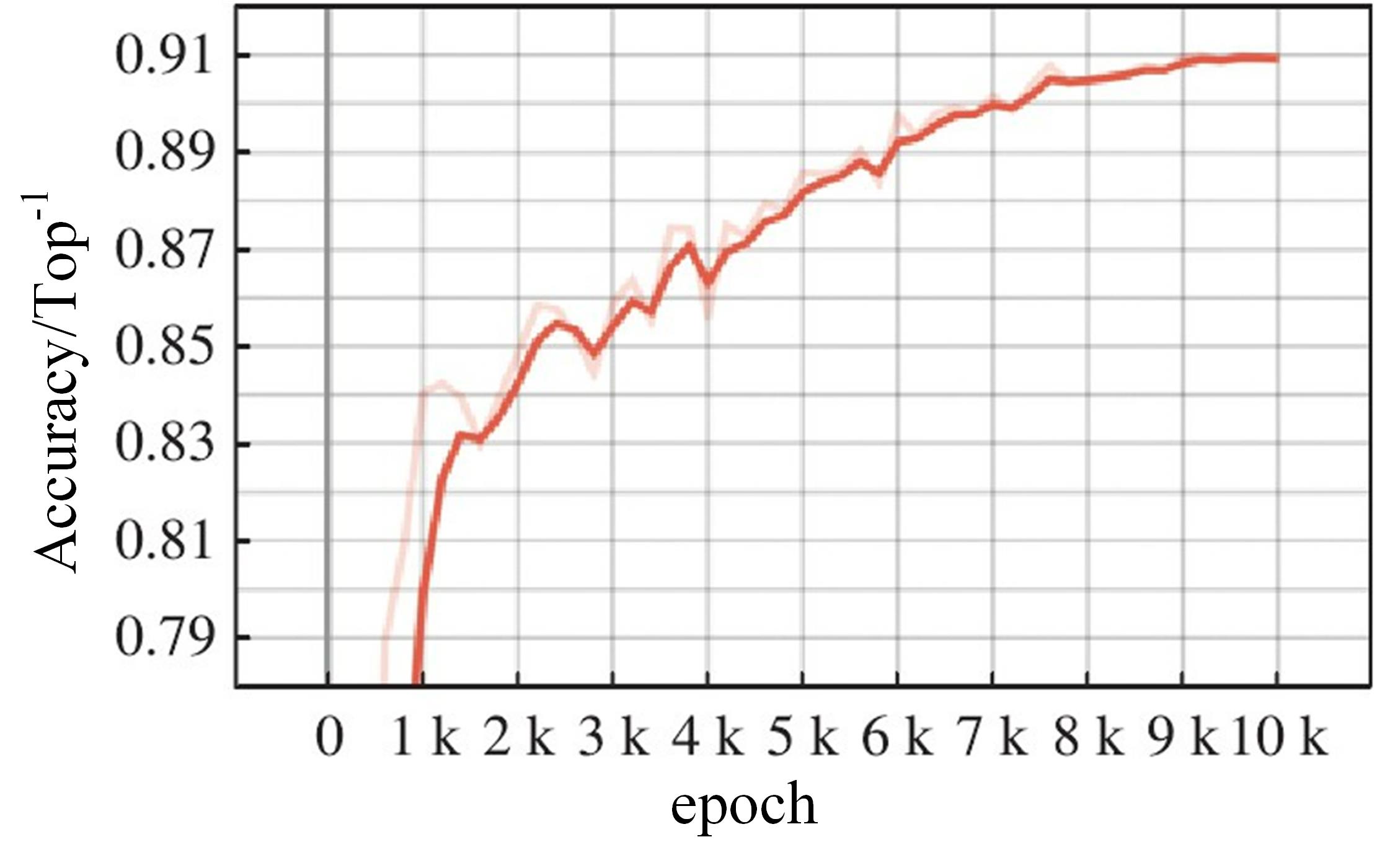}
            \caption{Stanford Dogs ass\label{fig4c}}
    \end{subfigure}
    \quad
    \begin{subfigure}[t]{0.225\textwidth}
            \centering
            \includegraphics[width=\textwidth]{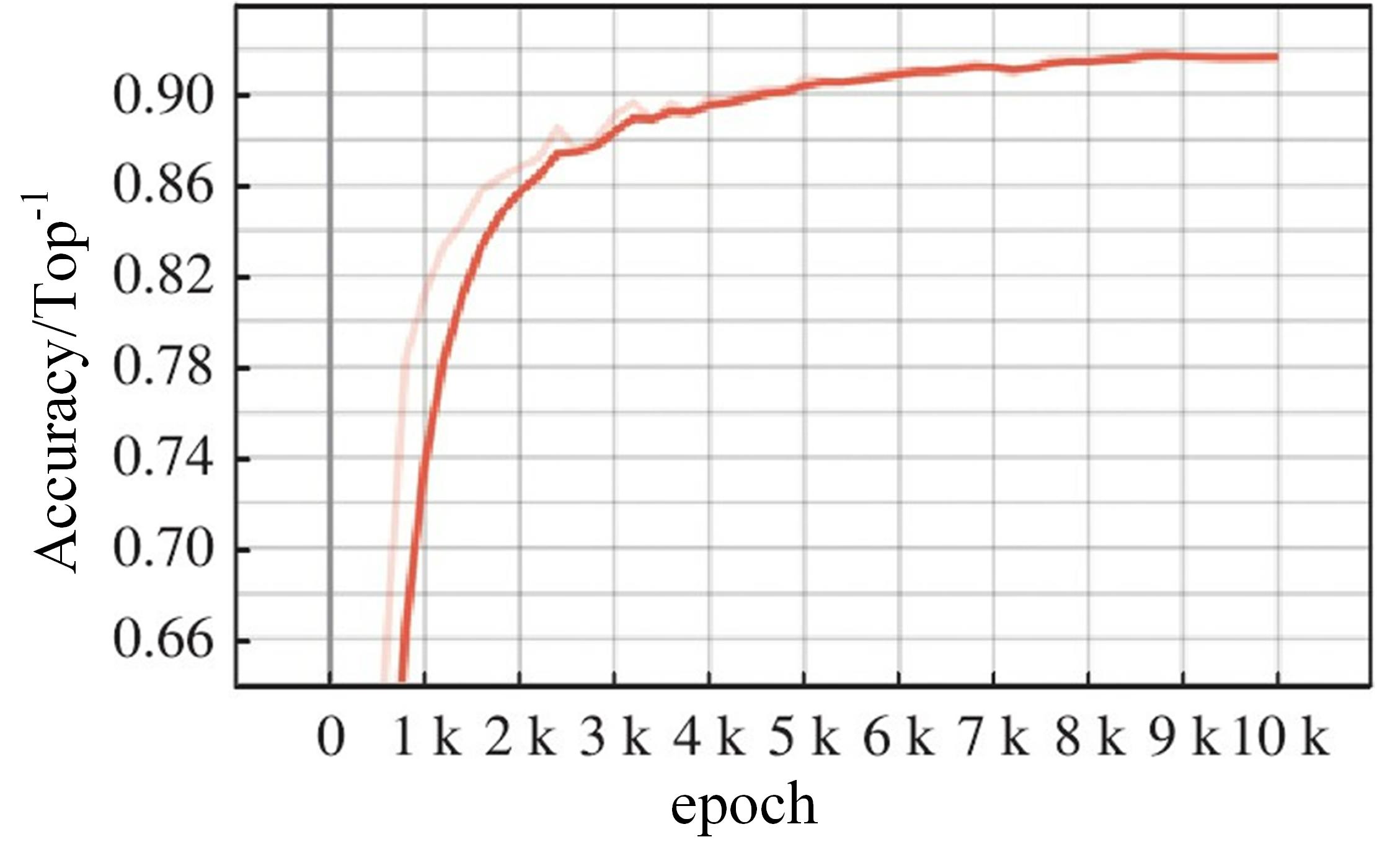}
            \caption{Stanford Dogs acc\label{fig4d}}
    \end{subfigure}
    \caption{Training loss and test accuracy under CUB-200-2011 and Stanford Dogs datasets\label{fig:4}}
\end{figure*}

\begin{figure}
	\centering
    \includegraphics[width=0.45\textwidth]{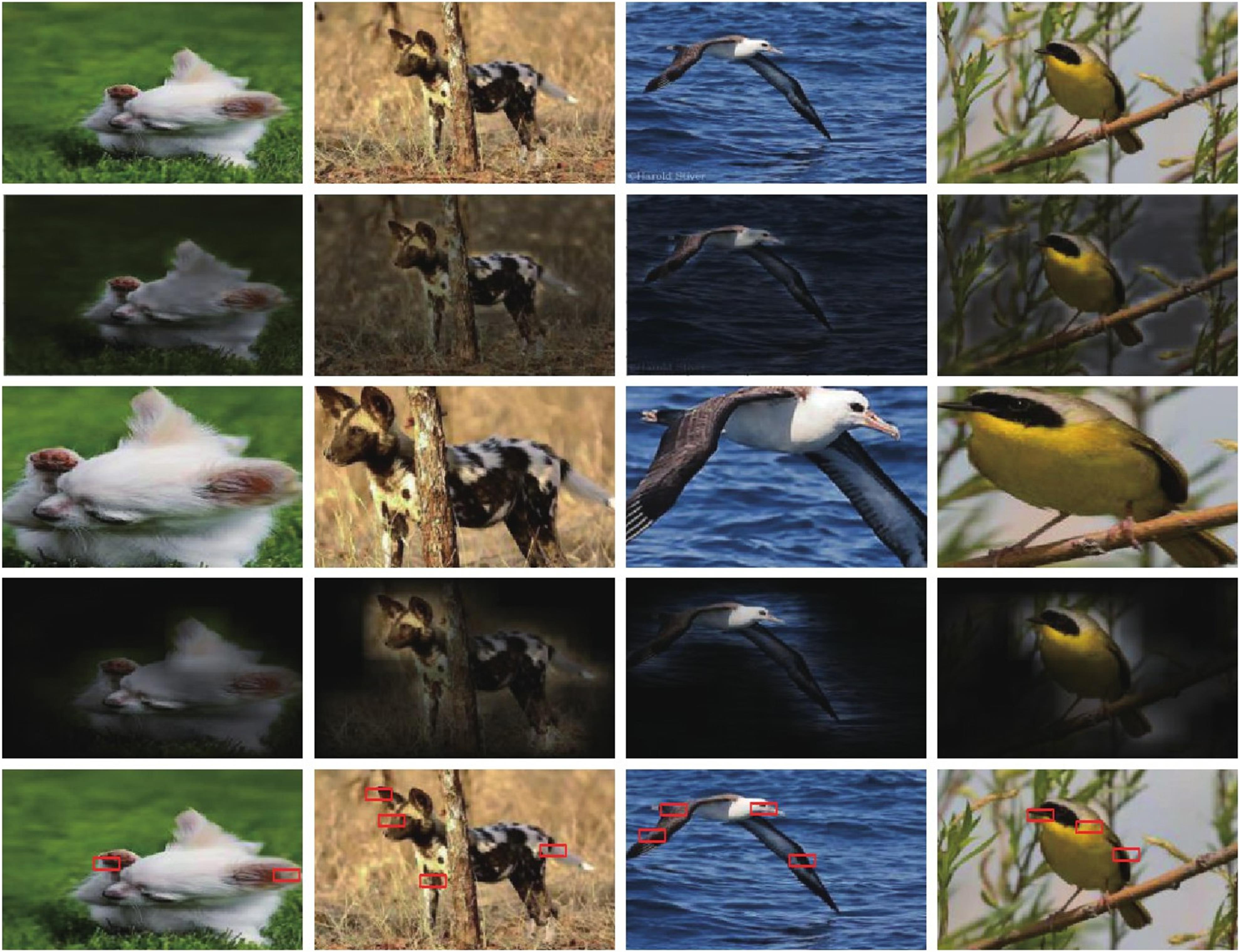}
	\caption{Results of attention visualization and data augmentation}
	\label{fig:5}
\end{figure}

\subsection{Experiment Results}
In this letter, an ablation study of DAVT was carried out on CUB-200-2011 and the Stanford Dogs dataset to analyze the effect of the proposed method on fine-grained classification accuracy. As shown in Table \ref{tab1}, hierarchical attention selection (HAS) improved the model's performance on the two datasets by 1.1\% and 0.7\%, respectively. Because HAS was guided by the attention weights between the levels, filtering out discriminating marks and discarding some unimportant ones, forcing the network to learn from critical parts. The model has improved by 0.3\% and 0.5\% accuracy by introducing the Data Augmentation module because the attention-clipped image removed some background noise. The data augmentation method introduces newly enhanced images while improving the classification accuracy, so the number of parameters and classification time will vary with the increase of training data during the training process.

Table \ref{tab2} shows the effect of data augmentation guided by different Transformer layer attention maps on the CUB-200-2011 dataset. $\xi$ indicates that the attention map of the first few transformer layers is used as a guide. It can be seen from Table \ref{tab2} that the effect is best when selecting the $5th$ layer attention map as the guide. The effect worsens as the number of layers gradually decreases or increases. When the number of layers is low or high, it is easy to cause the selected features to stay in the low or high-level local area. Therefore, select the Transformer layer in the middle part to extract the attention map to obtain more balanced feature information.

As shown in Table \ref{tab3}, the proposed method outperforms all methods with ViT and convolutional neural networks as the underlying network on the CUB-200-2011 and Stanford Dogs datasets. On the CUB-200- 2011 dataset, DAVT achieved a 0.2\% improvement on the Top-1\% metric compared to the best algorithm to date and a 1.4\% improvement compared to the original frame ViT \cite{ref5}. Compared to other mainstream CNNs, DAVT has significantly improved performance on both datasets. On the Stanford Dogs dataset, DAVT achieved a 0.4\% improvement in the Top-1\% metric compared to the best results to date and a 1.6\% improvement compared to the original framework ViT. In addition, to explore the effect of distance between different class sample feature representations on model performance, we compare the model's performance under different loss functions, as shown in Fig. \ref{fig:2}.

By evaluating the model on two standard FGVC datasets, the proposed method outperforms the existing ViT model-based network method compared to the existing Transformer classification method, as shown in Fig. \ref{fig:3}. Fig. \ref{fig:4} shows the loss and accuracy curves. The orange curve indicates the trend of loss and accuracy. Tensorboard generates the curve, and the light represents the real data curve. This letter obtains a dark curve by changing the smoothing factor to show the change in accuracy and loss better. As shown in Fig. \ref{fig4a} and \ref{fig4b}, the proposed method steadily reduces the training loss on both datasets. As a result of training the network using the weights of the pre-trained ViT-B\_16 model, the test accuracy improved rapidly over the first 2000 cycles, as shown in Fig. \ref{fig4c} and Fig. \ref{fig4d}. At the same time, the test accuracy curve has no downward trend, which proves that no fitting phenomenon occurs.

Fig. \ref{fig:5} shows the visual results of the model on two datasets. Specifically, row 1 of Fig. \ref{fig:5} shows the raw image of the dataset. Line 2 shows the attention map of the image, where light colors indicate the color of essential areas. Lines 3 and 4 of Fig. \ref{fig:5} show the data augmentation image. The dark areas in row 4 represent the background noise of attention-force cropping, as shown in Fig. \ref{fig:5}, where attention guide image cropping removes most of the background noise. Line 3 shows the cropped image and enlarges critical areas of the object, prompting the network to extract finer local features. Line 5 of Fig. \ref{fig:5} shows the location of the key markers for model positioning, where the red box represents the feature area of attention-emphasized, which the visualizer tools generated. As shown in Fig. \ref{fig:5}, the DAVT model successfully captured critical areas of the object, namely the head, wings, and tail of birds and the ears, eyes, and legs of dogs.

\section{Conclusion}
The input pixel blocks cause background noise in the original ViT model and the inability to efficiently extract subtle features between levels. In this letter, a data augmentation method for hierarchical attention selection and attention guidance is proposed, which reduces the interference of background noise on the network and forces the network to learn discriminant feature information between levels.


\end{document}